\documentclass[journal]{IEEEtran}
\usepackage{graphicx}
\usepackage{bm}
\usepackage{amssymb}
\usepackage{amsthm}
\usepackage{algorithm}
\usepackage{algorithmic}
\usepackage{lineno,hyperref}

\ifCLASSINFOpdf
\else
\fi
\newtheorem{theorem}{Theorem}
\newtheorem{lemma}{Lemma}

\newcommand{\eqn}[2]
{
	\begin{equation}#1
	#2
	\end{equation}
}

\newcommand{\eqnn}[1]
{
	\begin{displaymath}
	#1
	\end{displaymath}
}

\newcommand{\eqna}[1]
{
	\begin{eqnarray}
	#1
	\end{eqnarray}
}

\newcommand{\FR}
{ \FOR}

\newcommand{\EFR}
{ \ENDFOR}

\newcommand{\ST}
{\STATE}

\newcommand{\EIF}
{\ENDIF}

\newcommand{\thm}[1]
{
	\begin{theorem}
		#1
	\end{theorem}
}

\newcommand{\lem}[1]
{
	\begin{lemma}
		#1
	\end{lemma}
}

\newcommand{\prf}[1]
{ \begin{proof}
		#1
\end{proof}}

\newcommand{\tbl}[4]
{
	\begin{table}[htbp]
		\centering
		\caption{#2}
		#1
		\begin{tabular}{#3}
			#4
		\end{tabular}
	\end{table}
}

\newcommand{\fgr}[4]
{
	\begin{figure*}[!htbp]
		\centering
		\includegraphics[angle=0, width=#3\textwidth]{#4}
		\caption{#2}#1
	\end{figure*}
}

\newcommand{\sn}[1]
{
	\section{#1}
}

\newcommand{\ssn}[1]
{
	\subsection{#1}
}

\newcommand{\tf}[1]
{
	\textbf{#1}
}

\newcommand{\alg}[4]
{
	\begin{algorithm}[!htp]
		\caption{#2}#1
		\begin{algorithmic}
			#3
		\end{algorithmic}
		#4
	\end{algorithm}
}

\newcommand{\nn}
{
	\nonumber
}

\newcommand{\ra}
{
	\rightarrow
}

\begin{document}
	%
	\title{Stochastic Conjugate Gradient Algorithm with Variance Reduction}
	%
	%
	%
	
	\author{Xiao-Bo~Jin, Xu-Yao~Zhang, Kaizhu~Huang and Guang-Gang~Geng
		\thanks{Xiao-Bo Jin is with the Department
			of Information Science and Engineering, Henan University of Technology, Zhengzhou,
			Henan, China, 450001, email: xbjin9801@gmail.com; Xu-Yao Zhang is an Associate Professor at the National
			Laboratory of Pattern Recognition (NLPR), Institute
			of Automation, Chinese Academy of Sciences, Beijing, China; Kaizhu Huang works as the Head of the Department
			of Electrical and Electronic Engineering at Xian
			Jiaotong-Liverpool University; Guang-Gang Geng
			is with the Computer Network Information Center,
			Chinese Academy of Sciences, Beijing.}}

	%
	%

	\markboth{IEEE TRANSACTIONS ON NEURAL NETWORKS AND LEARNING SYSTEMS}%
	{Shell \MakeLowercase{\textit{et al.}}: Bare Demo of IEEEtran.cls for IEEE Journals}
	%

	
	
	\maketitle
	
	\begin{abstract}
		Conjugate gradient methods are a class of important methods for solving linear equations and nonlinear optimization problems. In this work, we propose a new stochastic conjugate gradient algorithm with variance reduction (CGVR \footnote{CGVR algorithm is available on github: https://github.com/xbjin/cgvr}), and we prove its linear convergence with the Fletcher and Reeves method for strongly convex and smooth functions. We experimentally demonstrate that the CGVR algorithm converges faster than its counterparts for four learning models, which may be convex, nonconvex or nonsmooth. Additionally, its area under the curve (AUC) performance on six large-scale datasets is comparable to that of the LIBLINEAR solver for the $L2$-regularized $L2$-loss but with a significant improvement in computational efficiency.
	\end{abstract}


	%
		
	\sn{Introduction}
	Empirical risk minimization (ERM) is a principle in statistical learning theory for providing theoretical bounds on the performance of learning algorithms. ERM is defined as
	\eqn{\label{eqn:erm_loss}}{
		\min_{\bm{w}}f(\bm{w}) = \frac{1}{n}\sum_{i = 1}^{n} f_i(\bm{w}),
	}
	where $\bm{w} \in R^d$ is the parameter of a machine learning model, $n$ is the sample size, and each $f_i(\bm{w}): R^d \ra R$ estimates how well parameter $\bm{w}$ fits the data of the i-th sample. This approach has been widely used to solve classification \cite{jin_regularized_2010}, regression \cite{zhang_retargeted_2015}, clustering \cite{jin_accelerating_2018} and ranking \cite{jin_combination_2015,jin_approximately_2018}, among others.
	
	Stochastic gradient descent (SGD) \cite{bottou_large-scale_2010} and its variants \cite{dozat_incorporating_2016,reddi_convergence_2018} are the most widely used algorithms for minimizing the empirical risk (\ref{eqn:erm_loss}) in many large-scale machine learning problems. Kingma and Diederik \cite{kingma_adam:_2015} introduced  the Adam method, which computes the adaptive learning rates for each parameter. Sutskever et al. \cite{sutskever_importance_2013} showed that SGD with momentum, using a well-designed random initialization and a slowly increasing schedule for the momentum parameter, could train both DNNs and RNNs. However, the success of these SGD variants heavily relies on the setting of the initial learning rate and the decay strategy of the learning rate.
	
	The disadvantage of SGD is that the randomness introduces a variance, which slows the convergence. Le Roux et al. \cite{roux_stochastic_2012} proposed stochastic average gradient (SAG) to achieve a variance reduction effect for SGD. Shalev-Shwartz and Zhang \cite{shalev-shwartz_stochastic_2013} introduced stochastic dual coordinate ascent (SDCA) to train convex linear prediction problems with a linear convergence rate. However, both methods require storing all gradients (or dual variables), thus making them unsuitable for complex applications where storing all gradients is impractical. The stochastic variance reduced gradient (SVRG) method proposed by Johnson and Zhang \cite{johnson_accelerating_2013} accelerates the convergence of stochastic first-order methods by reducing the variance of the gradient estimates.
	
	Another promising line of work is devoted to the stochasticization of the second-order quasi-Newton method, particularly the L-BFGS algorithm. Wang et al. \cite{wang_stochastic_2014} studied stochastic quasi-Newton methods for nonconvex stochastic optimization. Mokhtari and Ribeiro \cite{mokhtari_res:_2014} used stochastic gradients in lieu of deterministic gradients to determine the descent directions and approximate the objective function's curvature. Moritz et al. \cite{moritz_linearly-convergent_2016} introduced a stochastic variant of L-BFGS (SLBFGS) that incorporated the idea of variance reduction. Gower et al. \cite{gower_stochastic_2016} proposed a new limited-memory stochastic block BFGS update with the variance reduction approach SVRG. However, the limited-memory stochastic quasi-Newton methods often require $m$ vector pairs to efficiently compute product $H\nabla f$ ($H$ is the Hessian), which may be prohibitive in the case of limited memory for large-scale machine learning problems. 
	
	Fletcher and Reeves \cite{fletcher_function_1964} first showed how to extend the linear conjugate gradient (CG) method to nonlinear functions, which is called the FR method. Polak and Ribiere \cite{polak_note_1969} proposed another CG method known as the PR method. Gilbert and Nocedal proved that the modified PR method $\beta_k^{PR+} = \max\{0,\beta_k^{PR}\}$ with Wolfe-Powell linear search is globally convergent under a sufficient descent condition. In practical computation, the PR method, HS \cite{hestenes_methods_1952} method, and LS \cite{liu_efficient_1991} method are generally believed to be the most efficient CG methods because they essentially restart if a bad direction occurs. Although the convergence of the CD \cite{dennis_practical_1982} method, DY \cite{dai_nonlinear_1999} method, and FR method has been established, their numerical results are not good.

	In this work, we propose a stochastic variant of the CG method called CGVR, which integrates the variance reduction method. The proposed method has the following advantages: (1) It only requires a few iterations to quickly converge because of the idea of SVRG, but it converges more quickly than SVRG because of the use of the CG rather than the general gradient. (2) The parameters of CGVR are not sensitive to the datasets, and the empirical settings always work well; in particular, its step size is determined through a Wolfe line search. (3) It only stores the last gradient vector similar to CG; in contrast, the quasi-Newton variants often store a set of vector pairs. (4) CGVR with $L2$-regularized $L2$-loss achieves a generalization performance comparable to that of the LIBLINEAR solver \cite{fan_liblinear:_2008} in less running time for large-scale machine learning problems. 
	
	Our \tf{contributions} are as follows:
	
	(1) We propose a stochastic variant of the CG method with variance reduction, where both Wolfe line search and gradient computation are built on subsamples. 
	
	(2) We prove the linear convergence of CGVR with the Fletcher and Reeves method for strongly convex and smooth functions. 
	
	(3) We conduct a series of experiments on six large-scale datasets with four state-of-the-art learning models, which may be convex, nonconvex, or nonsmooth. The experimental results show that CGVR converges faster on large-scale datasets than several other algorithms, and its area under the curve (AUC) performance with $L2$-regularized $L2$-loss is comparable to that of the LIBLINEAR solver with a significant improvement in computational efficiency. 
	
	The remainder of this paper is organized as follows. In Section 2, we briefly introduce the SVRG and SLBFGS algorithms. In Section 3, we propose our CGVR algorithm and prove its linear convergence for strongly convex and smooth functions. In Section 4, we conduct experiments on convergence and generalization to compare CGVR with its counterparts. Finally, we draw some conclusions in Section 5.

	\sn{SVRG and SLBFGS Algorithms}
	\ssn{SVRG Algorithm}
	The SVRG algorithm was proposed by Johnson and Zhang \cite{johnson_accelerating_2013} for optimizing (\ref{eqn:erm_loss}) and is depicted in Alg. \ref{alg:svrg}.
	
	\alg{\label{alg:svrg}}{Stochastic Variance Reduced Gradient}{
		\ST Given $\bm{w}_0$, update frequency $m$, step size $\alpha$
		\FR{$k = 0,1,\cdots,T$}
		\ST $\bm{u}_k =  \nabla f (\bm{w}_k)$
		\ST $\bm{x}_0 = \bm{w}_k$
		\FR{$t = 0,1,\cdots,m - 1$}
		\ST Randomly select $i_t \in \{1,2,\cdots,n\}$
		\ST $\bm{g}_t = \nabla f_{i_t}(\bm{x}_t) - \nabla f_{i_t}(\bm{x}_0) + \bm{u}_k$
		\ST $\bm{x}_{t + 1} = \bm{x}_t - \alpha \bm{g}_t$
		\EFR
		\ST \tf{Option I: } $\bm{w}_{k + 1} = \bm{x}_m$
		\ST \tf{Option II: } $\bm{w}_{k + 1}$ for randomly chosen $t \in \{1,2,\cdots,m\}$
		\EFR
	}{}
	
	There are two loops in Alg. \ref{alg:svrg}. In the outer loop, the full gradient $\bm{u}_k$ is computed. We retain a snapshot $\bm{x}_0$ of $\bm{w}$ after every $m$ SGD iterations. In the inner loop, we randomly select an example from the dataset $X$ to produce a variance-reduced gradient estimate (see the proof of Theorem 1 in \cite{johnson_accelerating_2013}). There are two options to select the next $\bm{w}$ (e.g., $\bm{w}_{k + 1}$). Although Option I is a better choice than Option II because it takes more iterations to obtain the next $\bm{w}$, the convergence analysis is only available for Option II \cite{johnson_accelerating_2013}.
	
	\ssn{SLBFGS}
	For a subset $\mathcal{S} \subset \{1,2,\cdots, n\}$, we define the subsampled function $f_{\mathcal{S}}(\bm{w})$ as
	\eqn{\label{eqn:sample_gradient}}{
		f_{\mathcal{S}}(\bm{w}) = \frac{1}{|\mathcal{S}|} \sum_{i \in \mathcal{S}} f_{i}(\bm{w}),
	}
	where $|\mathcal{S}|$ denotes the number of elements in the set $\mathcal{S}$. Correspondingly, our algorithm uses the stochastic estimates of the gradient $\nabla f_{\mathcal{S}}$. In addition, we use stochastic approximations for the inverse Hessian $\nabla^2 f_{\mathcal{T}}$, where $\mathcal{T} \subset \{1,2,\cdots,n\}$ is different from $\mathcal{S}$ to decouple the estimation of the gradient from the estimation of the Hessian.
	
	The algorithm performs a full gradient computation every $m$ iterations and updates the inverse Hessian approximation every $L$ iterations. The vector $\bm{y}_r$ is computed by the product of the stochastic approximation of the Hessian and the vector $\bm{s}_r$, where $\bm{s}_r$ is the difference of two consecutive sequences with length $L$. The product $H_r \bm{g}_t$ is directly obtained from the two-loop recursion, whose inputs are the most recent $M$ vector pairs $\{(\bm{s}_j,\bm{y}_j)\}_{j = r - M + 1}^r$.
	
	\alg{\label{alg:s-lbfgs}}{Stochastic L-BFGS}{
		\ST Initialize $r = 0$
		\ST $H_0 = I$
		\FR{$k = 0,1,\cdots,T$}
		\ST $\bm{u}_k = \nabla f(\bm{w}_k)$
		\ST $\bm{x}_0 = \bm{w}_k$
		\FR{$t = 0,1,\cdots,m - 1$}
		\ST Sample a minibatch $\mathcal{S}_{k,t} \subset \{1,2,\cdots,n\}$
		\ST $\bm{g}_t = \nabla f_{\mathcal{S}_{k,t}}(\bm{x}_t) - \nabla f_{\mathcal{S}_{k,t}}(\bm{x}_0) + \bm{u}_k$
		\ST $\bm{x}_{t + 1} = \bm{x}_t - \alpha H_r \bm{g}_t$
		
		\IF{$t$ mod $L == 0$}
		\ST $r = r + 1$
		\ST $\bm{v}_r = \frac{1}{L}\sum_{j = t - L}^{t - 1}\bm{x}_j$
		\ST Sample a minibatch $\mathcal{T}_r \subset \{1,2,\cdots,n\}$
		\ST $\bm{s}_r = \bm{v}_r - \bm{v}_{r - 1}$
		\ST $\bm{y}_r = \nabla^2 f_{\mathcal{T}_r}(\bm{v}_r)\bm{s}_r$ 
		\ST Compute $H_r \bm{g}_t$ with $\bm{g}_t$ and $\{(\bm{s}_j,\bm{y}_j)\}_{j = r - M + 1}^r$ by two-loop recursion  
		\EIF
		\EFR
		\ST \tf{Option I: } $\bm{w}_{k + 1} = \bm{x}_m$
		\ST \tf{Option II: } $\bm{w}_{k + 1}$ for randomly chosen $t \in \{1,2,\cdots,m\}$  
		\EFR
	}{}

	\sn{Stochastic Conjugate Gradient with Variance Reduction}
	Although SVRG accelerates the convergence of SGD by reducing the variance of the gradient estimates, it is sensitive to the learning rate. SLBFGS requires  $M$ vector pairs to compute the product $H \nabla f$,  and it needs to calculate the Hessian matrix $H$.
	In the following, we propose a new algorithm called stochastic conjugate gradient with variance reduction (CGVR) to overcome the above disadvantages.
	
	\ssn{Framework of Algorithm}
	We adapt the CG algorithm \cite{nocedal_numerical_2006} from SVRG to obtain the CGVR algorithm in Alg. \ref{alg:s-cgvr}. We compute a variance-reduced gradient $\bm{g}_{t + 1}$ on the set $\mathcal{S}_{k,t} \subset \{1,2,\cdots,n\}$, which is randomly generated in the t-th loop of the k-th iteration:
	\eqn{}{
		\bm{g}_{t + 1} = \nabla f_{\mathcal{S}_{k,t}}(\bm{x}_{t + 1}) - \nabla f_{\mathcal{S}_{k,t}}(\bm{x}_0) + \bm{u}_k,
	} 
	where $\bm{g}_{t + 1}$ corresponds to $\nabla f(\bm{x}_{t + 1})$ in the CG algorithm, and $\nabla f_{\mathcal{S}_{k,t}}(\cdot)$ is computed by (\ref{eqn:sample_gradient}).
	
	Polak and Ribiere is a popular and important method in CG, which defines parameter $\beta_{t + 1}$ as follows:
	\eqn{}{
		\beta_{t + 1}^{PR} = \frac{\bm{g}_{t + 1}^T(\bm{g}_{t+1} - \bm{g}_{t})}{\bm{g}_{t}^T \bm{g}_{t}}.
	} 
	Simultaneously, the Fletcher-Reeves method uses another approach to compute $\beta_{t + 1}$:
	\eqn{\label{eqn:fr-update}}{
		\beta_{t + 1}^{FR} = \frac{\|\bm{g}_{t + 1}\|^2}{\|\bm{g}_t\|^2}.
	}
	
	We use a trick to set $\beta_{t + 1} = 0$ and restart \cite{powell_restart_1977} the iteration with the steepest descent step at the beginning of each iteration. Restarting will periodically refresh the algorithm and erase old information that may not be beneficial. Nocedal and J. Wright \cite{nocedal_numerical_2006} (see Equation (5.51) on Page 124) provide a strong theoretical result about restarting:
	It leads to $m$-step quadratic convergence, that is,
	\eqn{}{
		\|\bm{x}_{t + m} - \bm{x}^{*}\| = O(\|\bm{x}_t - \bm{x}^{*}\|^2),
	}
	where $\bm{x}^{*}$ is a local minimizer of the function.
	
	The search direction $\bm{p}_t$ may fail to be a descent direction unless $\alpha_t$ satisfies certain conditions. We can avoid this situation by requiring the step length $\alpha_t$ to satisfy the strong Wolfe conditions, which are
	\eqna{
		f(\bm{x}_t + \alpha_t \bm{p}_t) &\le& f(\bm{x}_t) + c_1 \alpha_t \nabla f(\bm{x}_t)^T \bm{p}_t \label{eqn:wolfe11},\\
		|\nabla f(\bm{x}_t + \alpha_t \bm{p}_t)^T \bm{p}_t| &\le& - c_2 \nabla f(\bm{x}_t)^T \bm{p}_t \label{eqn:wolfe12},
	}
	where $0 < c_1 < c_2 < 1$. In our experiments, we computed parameter $\beta$ as
	\eqn{}{
		\beta_{t + 1}^{PR+} = \max\{\beta_{t + 1}^{PR},0\},
	}
	which leads to the $PR^{+}$ method; then, the strong Wolfe condition ensures that the descent property holds.
	
	Note that 
	\eqn{\label{eqn:gradientf}}{
		\nabla f(\bm{x}_t) = \mathbb{E}[\bm{g}_t].
	}
	CGVR uses $f_{\mathcal{S}_{k,t}}(\bm{x}_t + \alpha \bm{p}_t)$ rather than $f(\bm{x}_t + \alpha \bm{p}_t)$ to search for the steps that satisfy the following conditions:
	\eqna{
		f_{\mathcal{S}_{k,t}}(\bm{x}_t + \alpha_t \bm{p}_t) &\le& f_{\mathcal{S}_{k,t}}(\bm{x}_t) \nn \\ 
		&& + c_1 \alpha_t \nabla f_{\mathcal{S}_{k,t}}(\bm{x}_t)^T \bm{p}_t \label{eqn:wolfe21},\\
		|\nabla f_{\mathcal{S}_{k,t}}(\bm{x}_t + \alpha_t \bm{p}_t)^T \bm{p}_t| &\le& - c_2 \nabla_{\mathcal{S}_{k,t}} f(\bm{x}_t)^T \bm{p}_t \label{eqn:wolfe22}.
	} 
	Although $f(\bm{x}_t + \alpha \bm{p}_t)$ is also possible, using $f_{\mathcal{S}_{k,t}}(\bm{x}_t + \alpha \bm{p}_t)$ for the linear search will clearly be faster. The values $c_1 = 10^{-4}$ and $c_2 = 0.1$ are commonly used in the CG algorithm. The initial search step is set to $1$.
	
	Since $\bm{g}_t$ in the CGVR algorithm replaces the role of $\nabla f(\bm{x}_t)$ in the classical CG algorithm, to borrow some conclusions of the CG algorithm, we also use the following condition in the convergence analysis:
	\eqn{\label{eqn:curve_condition}}{
		|\bm{g}_{t + 1}^T \bm{p}_t| \le - c_2 \bm{g}_t^T \bm{p}_t \label{eqn:wolfe_3}.
	}
	
	\alg{\label{alg:s-cgvr}}{Stochastic Conjugate Gradient with Variance Reduction}{
		\ST Given $\bm{w}_0$ 
		\ST $\bm{h}_0 = \nabla f(\bm{w}_{0})$
		
		\FR {$k = 0,1,\cdots, T - 1$}
		\ST $\bm{u}_k = \nabla f(\bm{w}_{k})$ 
		\ST $\bm{x}_0 = \bm{w}_k$
		\ST $\bm{g}_0 = \bm{h}_{k}$
		\ST $\bm{p}_0 = - \bm{g}_0$
		\FR{$t = 0,\cdots,m - 1$}
		\ST Sample a minibatch $\mathcal{S}_{k,t} \subset \{1,2,\cdots,N\}$
		\ST Call the line search algorithm to find $\alpha_t$ approximately optimize:
		\eqnn{
			\min_{\alpha}f_{\mathcal{S}_{k,t}}(\bm{x}_t + \alpha \bm{p}_t).
		}
		\ST $\bm{x}_{t + 1} = \bm{x}_t + \alpha_t \bm{p}_t$
		
		\ST Compute $\bm{g}_{t + 1} = \nabla f_{\mathcal{S}_{k,t}}(\bm{x}_{t + 1}) - \nabla f_{\mathcal{S}_{k,t}}(\bm{x}_0) + \bm{u}_k$
		
		\ST Compute $\beta$ by 
		\ST \tf{Option I: } 
		\eqnn{
			\beta^{PR+}_{t + 1} = \max \left\{\frac{\bm{g}_{t + 1}^T(\bm{g}_{t+1} - \bm{g}_{t})}{\bm{g}_{t}^T \bm{g}_{t}},0\right\}
		}
		\ST \tf{Option II: }
		\eqnn{
			\beta_{t + 1}^{FR} = \frac{\|\bm{g}_{t + 1}\|^2}{\|\bm{g}_t\|^2}
		}
		
		\ST $\bm{p}_{t + 1} = - \bm{g}_{t + 1} + \beta_{t + 1} \bm{p}_t$   
		\EFR
		
		\ST $\bm{h}_{k + 1} = \bm{g}_m$
		
		\ST \tf{Option I: } $\bm{w}_{k + 1} = \bm{x}_m$
		\ST \tf{Option II: } $\bm{w}_{k + 1} = \bm{x}_t$ for randomly chosen $t \in \{0,1,\cdots,m - 1\}$
		
		\EFR   
	}{}

	The ideal step length would be the global minimizer of the univariate function $\phi(\cdot)$ defined by
	\eqn{}{
		\phi(\alpha) = f_{\mathcal{S}_{k,t}}(\bm{x}_t + \alpha \bm{p}_t).
	}
	We can perform an inexact line search to identify a step length. The line search algorithm \cite{nocedal_numerical_2006} is divided into two steps: the first step begins from an initial estimate $\alpha_1$ and constantly increases this estimate until it finds a suitable step length or a range that includes the desired step length; the second step is invoked by calling a function called $zoom$, which successively decreases the size of the interval until an acceptable step length is identified.
	
	We stop the line search and $\bm{zoom}$ procedure if it cannot obtain a lower function value after 20 iterations. In the $\bm{zoom}$ procedure, we use the middle point of the interval as a new candidate rather than complex quadratic, cubic, or bisection interpolation. These tricks work well in practice.

	Finally, we use Option I to obtain the next $\bm{w}$, but the convergence analysis is only available for Option II.
	
	\ssn{Convergence Analysis}
	For the convenience of discussion, we define
	\eqn{}{
		\delta_f(\bm{x})  =  f(\bm{x}) - f(\bm{w}_{*}),\quad f_t = f(\bm{x}_t), \quad \nabla f_t = \nabla f(\bm{x}_t).
	} 
	
	We now investigate the convergence of the CGVR algorithm by updating $\beta_t$ with Fletcher-Reeves update (\ref{eqn:fr-update}) (Option II).
	
	In the following discussion, we use $\beta_t$ to represent $\beta_t^{FR}$ unless otherwise specified. Our analysis uses the following assumptions.
	
	\tf{Assumption 1. } The CGVR algorithm is implemented with a step length $\alpha_t$ that satisfies $\alpha_t \in [\alpha_{lo},\alpha_{hi}]$ ($0 < \alpha_{lo} < \alpha_{hi}$) and condition (\ref{eqn:wolfe_3}) with $c_2 < 1/5$.
	
	\tf{Assumption 2. } The function $f_i$ is twice continuously differentiable for each $1 \le i \le n$, and there exist constants $0 < \lambda \le \Lambda$ such that  
	\eqn{}{
		\lambda I \preceq \nabla^2 f_{\mathcal{S}}(\bm{x}) \preceq \Lambda I
	}
	for all $\bm{x} \in \mathbb{R}^d$ and all $\mathcal{S} \subset \{1,2,\cdots,N\}$.
	
	\tf{Assumption 3. }  There exists $\hat{\beta} < 1$ such that 
	\eqn{}{
		\beta_{t + 1} = \frac{\|\bm{g}_{t + 1}\|^2}{\|\bm{g}_t\|^2} \le \hat{\beta}.
	}
	
	In the following Lemmas \ref{lem:delta_f} and \ref{lem:v_t_upper_bound}, we estimate the lower bound of $\|\nabla f(\bm{x})\|$ and the upper bound of $\mathbb{E}[\|\bm{g}_{t}\|^2]$, whose proofs are provided in Lemmas 5 and 6 of \cite{moritz_linearly-convergent_2016}. 
	
	\lem{\label{lem:delta_f} Suppose that $f$ is continuously differentiable and strongly convex with parameter $\lambda$. Let $\bm{w}_{*}$ be the unique minimizer of $f$. Then, for any $\bm{x} \in \mathbb{R}^d$, we have
		\eqn{}{
			\|\nabla f(\bm{x})\|^2 \ge 2 \lambda \delta_f(\bm{x}).
		}
	}
	
	\lem{\label{lem:v_t_upper_bound} Let $\bm{w}_{*}$ be the unique minimizer of $f$. Let $\bm{u}_k = \nabla f(\bm{w}_k)$, and let $\bm{g}_{t} = \nabla f_{\mathcal{S}_{k,t}}(\bm{x}_{t}) - \nabla f_{\mathcal{S}_{k,t}}(\bm{w}_k) + \bm{u}_k$ be the variance-reduced stochastic gradient. Taking an expectation with respect to $\mathcal{S}_{k,t}$, we obtain 
		\eqn{}{
			\mathbb{E}[\|\bm{g}_{t}\|^2] \le 4\Lambda (\delta_f(\bm{x}_t) + \delta_f(\bm{w}_k)).	
		}
	}
	
	The following Theorem \ref{thm:al-baali} is used to estimate the upper and lower bounds of $\bm{g}_t^T \bm{p}_t/\|\bm{g}_t\|^2$, which is proven using the mathematical induction method; see Lemma 5.6 in \cite{nocedal_numerical_2006} for details.
	
	\thm{\label{thm:al-baali} Suppose that the CGVR (or CG) algorithm is implemented with step length $\alpha_t$ that satisfies condition (\ref{eqn:wolfe_3}) with $0 < c_2 < 1/2$; then, the FR method \cite{al-baali_descent_1985,nocedal_numerical_2006} generates descent directions $\bm{p}_k$ that satisfy
		\eqn{\label{eqn:gp_equality}}{
			- \frac{1}{1 - c_2} \le \frac{\bm{g}_t^T \bm{p}_t}{\|\bm{g}_t\|^2} \le \frac{2c_2 - 1}{1 - c_2}.
		}	
	}
	
	The following two theorems will state our main results.
	
	\thm{\label{thm:p_k_upper_bound} Suppose that Assumptions 1 and 3 hold for Alg. \ref{alg:s-cgvr}. Then, for any $t$, we have 
		\eqn{}{
			\mathbb{E}[\|\bm{p}_t\|^2] \le \eta(t)\mathbb{E}[\|\bm{g}_0\|^2],
		}
		where
		\eqn{}{
			\eta(t) = \frac{2}{1 - \hat{\beta}}\hat{\beta}^t - \frac{1 + \hat{\beta}}{1 - \hat{\beta}}\hat{\beta}^{2t}.
		}	
	}
	
	\prf{
		Combining condition (\ref{eqn:wolfe_3}) with Thm. (\ref{thm:al-baali}), we have
		\eqna{\label{eqn:f_tp_t}
			- \bm{g}_t^T \bm{p}_{t - 1} & \le & - c_2 \bm{g}_{t - 1}^T \bm{p}_{t - 1} \nn \\
			& \le & \frac{c_2}{1 - c_2}\|\bm{g}_{t - 1}\|^2.
		}
		
		According to Assumption 3, we obtain
		\eqn{\label{eqn:gt_expectation}}{
			\mathbb{E}[\|\bm{g}_{t}\|^2] \le \hat{\beta}\mathbb{E}[\|\bm{g}_{t - 1}\|^2].
		}
		
		Then, we use (\ref{eqn:f_tp_t}) and (\ref{eqn:gt_expectation}) to bound $\mathbb{E}[\|\bm{p}_t\|^2]$, \eqna{\
			\mathbb{E}[\|\bm{p}_t\|^2] 
			& = & \mathbb{E}[\|\beta_t
			\bm{p}_{t - 1} - \bm{g}_t\|^2] \nn \\
			& = & \mathbb{E}[\|\bm{g}_t\|^2] - 2 \beta_t E[\bm{g}_t^T \bm{p}_{t - 1}] + \beta_t^2 \mathbb{E}[\|\bm{p}_{t - 1}\|^2] \nn \\
			& \le & \hat{\beta}\mathbb{E}[\|\bm{g}_{t - 1}\|^2] + \frac{2 \hat{\beta} c_2}{1 - c_2}\mathbb{E}[\|\bm{g}_{t - 1}\|^2]\nn \\
			&& + \hat{\beta}^2 \mathbb{E}[\|\bm{p}_{t - 1}\|^2] \nn\\
			& = & \hat{\beta}\frac{1 + c_2}{1 - c_2}\mathbb{E}[\|\bm{g}_{t - 1}\|^2]
			+ \hat{\beta}^2 \mathbb{E}[\|\bm{p}_{t - 1}\|^2].
		}
		
		According to the monotonically increasing property of the function $(1 + x)/(1 -x)$ with $0 < x < 1$ and $c_2 < 1/5 < 1/3$, we can conclude that
		\eqn{}{
			\frac{1 + c_2}{1 - c_2} < \frac{1 + 1/3}{1 - 1/3} = 2.
		}
		Thus, we can immediately obtain the following inequality:
		\eqn{\label{eqn:p_t}}{
			\mathbb{E}[\|\bm{p}_{t}\|^2] \le 2\hat{\beta} \mathbb{E}[\|\bm{g}_{t - 1}\|^2] + \hat{\beta}^2 \mathbb{E}[\|\bm{p}_{t - 1}\|^2].
		}
		
		At the beginning of the k-th iteration, we have
		\eqn{\label{eqn:p0}}{
			\bm{p}_{0} = -\bm{g}_{0}. 
		}
		Furthermore, we unfold (\ref{eqn:gt_expectation}) until we reach $\bm{g}_{0}$:
		\eqn{\label{eqn:gt_seq}}{
			\mathbb{E}[\|\bm{g}_t\|] \le \hat{\beta}^{t} \mathbb{E}[\|\bm{g}_{0}\|].
		}
		
		According to (\ref{eqn:p_t}) and (\ref{eqn:gt_seq}), we further obtain
		\eqna{
			\mathbb{E}[\|\bm{p}_{t}\|^2] & \le & 2\hat{\beta} (\mathbb{E}[\|\bm{g}_{t - 1}\|^2] + \hat{\beta}^2\mathbb{E}[\|\bm{g}_{t - 2}\|^2] \nn \\
			&&
			+ \cdots + (\hat{\beta}^2)^{t - 1}\mathbb{E}[\|\bm{g}_{0}\|^2])
			+ (\hat{\beta}^2)^{t}\mathbb{E}[\|\bm{p}_{0}\|^2] \nn \\
			& = & 2\hat{\beta}(\hat{\beta}^{t - 1}\mathbb{E}[\|\bm{g}_{0}\|^2] + (\hat{\beta}^2) \hat{\beta}^{t - 2}\mathbb{E}[\|\bm{g}_{0}\|^2] \nn \\
			&& + \cdots + (\hat{\beta}^2)^{t - 1}\mathbb{E}[\|\bm{g}_{0}\|^2])  
			+ (\hat{\beta}^2)^{t}\mathbb{E}[\|\bm{g}_{0}\|^2] \nn \\
			& = & 2\hat{\beta} \mathbb{E}[\|\bm{g}_{0}\|^2] \sum_{j = 0}^{t - 1} (\hat{\beta}^2)^{j} \hat{\beta}^{t - 1 - j} + (\hat{\beta}^2)^{t}\mathbb{E}[\|\bm{g}_{0}\|^2] \nn \\
			& = & 2 \hat{\beta}^{t}\mathbb{E}[\|\bm{g}_{0}\|^2] \sum_{j = 0}^{t - 1} \hat{\beta}^j  + (\hat{\beta}^2)^{t}\mathbb{E}[\|\bm{g}_{0}\|^2] \nn \\
			& = & \left(2\hat{\beta}^{t}\frac{1 - \hat{\beta}^{t}}{1 - \hat{\beta}} + \hat{\beta}^{2t}\right) \mathbb{E}[\|\bm{g}_{0}\|^2] \nn \\
			& = & \frac{1}{1  - \hat{\beta}}(2\hat{\beta}^t - (\hat{\beta} + 1)\hat{\beta}^{2t})\mathbb{E}[\|\bm{g}_{0}\|^2]\nn\\
			& = & \eta(t)\mathbb{E}[\|\bm{g}_{0}\|^2] \label{eqn:pt},
		}
		where
		\eqn{\label{eqn:etat}}{
			\eta(t) = \frac{2}{1 - \hat{\beta}}\hat{\beta}^t - \frac{1 + \hat{\beta}}{1 - \hat{\beta}}\hat{\beta}^{2t}.
		}
	}
	
	\thm{\label{thm:s-cgvr}
		Suppose that Assumptions 1, 2 and 3 hold. Let $\bm{w}_{*}$ be the unique minimizer of $f$. Then, for all $k \ge 0$, we have
		\eqn{}{
			\mathbb{E}[f(\bm{w}_k) - f(\bm{w}_{*})] \le \xi^{k} \mathbb{E}[f(\bm{w}_0) - f(\bm{w}_{*})],
		}
		where the convergence rate is given by
		\eqn{}{
			\xi = \frac{1 + \hat{\beta}\alpha_{hi}\Lambda (m - 1) + 4\Lambda^2 \alpha_{hi}^2/(1 - \hat{\beta})^2}{(2 \alpha_{lo}\lambda  - \hat{\beta}\alpha_{hi}\Lambda) m + \hat{\beta}\alpha_{hi}\Lambda } < 1,
		}
		assuming that we choose
		$\alpha_{hi}/\alpha_{lo} < \lambda/(\hat{\beta}\Lambda)
		$
		and a sufficiently large $m$ to satisfy
		\eqn{}{
			m \ge \frac{1 - 2 \hat{\beta}\alpha_{hi}\Lambda + 4\Lambda^2 \alpha_{hi}^2/(1 - \hat{\beta})^2}{2\alpha_{lo}\lambda - 2\hat{\beta}\alpha_{hi}\Lambda}.
		}
		
	}
	
	\prf{
		Using the Lipschitz continuity of $\nabla f$ and Assumption 2, we have
		\eqn{\label{eqn:lipschitz}}{
			f(\bm{x}_{t + 1}) \le f(\bm{x}_t) + \nabla f(\bm{x}_t)^T(\bm{x}_{t + 1} - \bm{x}_t) + \frac{\Lambda}{2} \|\bm{x}_{t + 1} - \bm{x}_t\|^2.
		}
		
		Note that $\bm{p}_0 = - \bm{g}_0$; then, we have $\nabla f_0^T E[\bm{p}_0] = - \|\nabla f_0\|^2$.
		Since $\bm{p}_{t} = - \bm{g}_{t} + \beta_{t} \bm{p}_{t - 1} (t \ge 1)$ and the random variables $\bm{g}_t$ and $\bm{p}_{t - 1}$ are independent, with (\ref{eqn:gradientf}), (\ref{eqn:curve_condition}) and (\ref{eqn:gp_equality}), we have
		\eqna{
			\nabla f_t^T \mathbb{E}[\bm{p}_t] & = & \mathbb{E}[\bm{g}_t]^T \mathbb{E}[-\bm{g}_t + \beta_t \bm{p}_{t - 1}]	\nn \\
			& = & - \|\nabla f_t\|^2 + \beta_t \mathbb{E}[\bm{g}_t^T \bm{p}_{t - 1}]  \nn \\
			& \le & - \|\nabla f_t\|^2 - \beta_t c_2 \mathbb{E}[\bm{g}_{t - 1}^T \bm{p}_{t - 1}] \nn \\
			& \le & - \|\nabla f_t\|^2 + \beta_t \frac{c_2}{1 - c_2} \mathbb{E}[\|\bm{g}_{t - 1}\|^2] \nn \\
			& \le &  - \|\nabla f_t\|^2 + \frac{1}{4} \hat{\beta}\mathbb{E}[\|\bm{g}_{t - 1}\|^2].
		}
		The last expression is obtained from the monotonically increasing characteristic of the function $x/(1 - x)$. When $c_2 < 1/5$, \eqn{}{
			\frac{c_2}{1 - c_2} < \frac{1/5}{1 - 1/5} = \frac{1}{4}
		}
		
		Taking expectations on both sides of (\ref{eqn:lipschitz}), we obtain  	
		\eqna{\label{eqn:f_t_1}
			\mathbb{E}[f_{t + 1}] &\le& f_t
			+  \nabla f_t^T \mathbb{E}[(\bm{x}_{t + 1} - \bm{x}_t)] \nn\\
			&& + \frac{\Lambda}{2}\mathbb{E}[\|\bm{x}_{t + 1} - \bm{x}_t\|^2] \nn \\
			& \le & f_t
			+  \alpha_t (- \|\nabla f_t\|^2 + \frac{1}{4} \hat{\beta}\mathbb{E}[\|\bm{g}_{t - 1}\|^2]) + \nn \\
			&& \frac{\Lambda \alpha_t^2}{2}\mathbb{E}[\|\bm{p}_t\|^2] \nn \\
			& \le & f_t - \alpha_{lo} \|\nabla f_t\|^2 + \frac{\hat{\beta} \alpha_{hi}}{4} \mathbb{E}[\|\bm{g}_{t - 1}\|^2] \nn \\
			&&
			+ \frac{\Lambda \alpha_{hi}^2}{2}\eta(t)\cdot 4\Lambda (\delta_f(\bm{x}_0) + \delta_f(\bm{w}_k)) \nn \\
			& \le &  f_t - 2\alpha_{lo}\lambda \delta_f(\bm{x}_t) + \hat{\beta}\alpha_{hi}\Lambda (\delta_f(\bm{x}_{t - 1}) + \delta_f(\bm{w}_k)) \nn \\
			&&
			+ \tau \eta(t)  \delta_f(\bm{w}_k), 
		}
		where 
		\eqn{}{
			\tau = 4\Lambda^2 \alpha_{hi}^2.
		}
		
		Summing over $t = 0,1,\cdots, m - 1$ and using a telescoping sum, we obtain
		\eqna{\label{eqn:f_m}
			\mathbb{E}[f_m] &\le& \mathbb{E}[f_0]  - 2 \alpha_{lo}\lambda \left(\sum_{t = 0}^{m - 1}\mathbb{E}[\delta_f(\bm{x}_t)]\right) \nn \\
			&& + \hat{\beta}\alpha_{hi}\Lambda \sum_{t = 0}^{m - 2}(\mathbb{E}[\delta_f(\bm{x}_{t})]  + \mathbb{E}[\delta_f(\bm{w}_k)]) \nn \\
			&& + \tau \mathbb{E}[\delta_f(\bm{w}_k)]\sum_{t = 0}^{m - 1}\eta(t).
		}
		
		We now compute $\sum_{t = 0}^{m - 1}\eta(t)$,
		\eqna{
			\sum_{t = 0}^{m - 1}\eta(t) & = & \sum_{t = 0}^{m - 1}\frac{2}{1 - \hat{\beta}}\hat{\beta}^t - \frac{1 + \hat{\beta}}{1 - \hat{\beta}}(\hat{\beta}^2)^{t} \nn\\
			& = & \frac{2}{1 - \hat{\beta}}\frac{1 - \hat{\beta}^m}{1 - \hat{\beta}} - \frac{1 + \hat{\beta}}{1 - \hat{\beta}}\frac{1 - \hat{\beta}^{2m}}{1 - \hat{\beta}^2} \nn \\
			& = & \frac{(1 - \hat{\beta}^m)^2}{(1 - \hat{\beta})^2} \nn \\
			& \le & \frac{1}{(1 - \hat{\beta})^2}.
		}
		
		Rearranging (\ref{eqn:f_m}) provides
		\eqna{
			0 &\le& \mathbb{E}[f_0] - \mathbb{E}[f_m]  - 2 \alpha_{lo}\lambda m \mathbb{E}[\delta_f(\bm{w}_{k + 1})]\nn \\
			&& + \hat{\beta}\alpha_{hi}\Lambda (m - 1)\mathbb{E}[\delta_f(\bm{w}_{k + 1})] \nn \\
			&& + \hat{\beta}\alpha_{hi}\Lambda (m - 1)\mathbb{E}[\delta_f(\bm{w}_{k})] +  \tau/(1 - \hat{\beta})^2 \mathbb{E}[\delta_f(\bm{w}_k)] \nn \\
			& \le & \mathbb{E}[(f(\bm{w}_k) - f(\bm{w}_{*})] \nn \\
			&& +(\hat{\beta}\alpha_{hi}\Lambda (m - 1) -  2 \alpha_{lo}\lambda m)\mathbb{E}[\delta_f(\bm{w}_{k + 1})]\nn \\
			&& + (\hat{\beta}\alpha_{hi}\Lambda (m - 1) + \tau/(1 - \hat{\beta})^2)\mathbb{E}[\delta_f(\bm{w}_k)].
		}
		Furthermore, we have
		\eqn{}{
			\mathbb{E}[\delta_f(\bm{w}_{k + 1})] \le \xi \mathbb{E}[\delta_f(\bm{w}_k)],	
		}
		where
		\eqn{}{
			\xi = \frac{1 + \hat{\beta}\alpha_{hi}\Lambda (m - 1) + 4\Lambda^2 \alpha_{hi}^2/(1 - \hat{\beta})^2}{(2 \alpha_{lo}\lambda  - \hat{\beta}\alpha_{hi}\Lambda) m + \hat{\beta}\alpha_{hi}\Lambda }.
		}
		
		Let $\xi < 1$; then, it follows that
		\eqn{}{
			m \ge  \frac{1 - 2 \hat{\beta}\alpha_{hi}\Lambda + 4\Lambda^2 \alpha_{hi}^2/(1 - \hat{\beta})^2}{2\alpha_{lo}\lambda - 2\hat{\beta}\alpha_{hi}\Lambda}.
		}
		
		We observe that for $\hat{\beta} < 1$, we have
		\eqna{
			&& 1 - 2 \hat{\beta}\alpha_{hi}\Lambda + 4\Lambda^2 \alpha_{hi}^2/(1 - \hat{\beta})^2 \nn \\
			& = &(1 - \hat{\beta}\alpha_{hi}\Lambda)^2 + \left(\frac{4}{(1 - \hat{\beta})^2} - \hat{\beta}^2\right)\cdot \Lambda^2 \alpha_{hi}^2 \nn \\
			& > & 0 \nn.
		}
		Assuming that we choose the step interval $[\alpha_{lo},\alpha_{hi}]$ that satisfies 
		\eqn{}{
			\frac{\alpha_{hi}}{\alpha_{lo}} < \frac{\lambda}{\hat{\beta}\Lambda},
		}
		then a sufficiently large $m$ will ensure the linear convergence of CGVR.
	}
	
	\sn{Experiments}
	
	In this section, we compare our algorithm CGVR with SGD, SVRG \cite{johnson_accelerating_2013}, CG  \cite{nocedal_numerical_2006}, and SLBFGS \cite{moritz_linearly-convergent_2016}. Our experiments show the effectiveness of CGVR on several popular learning models, which may be convex, nonconvex or nonsmooth. 
	
	\ssn{Descriptions of Models and Datasets}
	We evaluate these algorithms on four state-of-the-art learning models:
	
	(1) ridge regression (ridge)
	\eqn{\label{eqn:ridge}}{
		\min_{\bm{w}} \frac{1}{n}\sum_{i = 1}^n (y_i - \bm{x}_i^T \bm{w})^2 + \lambda \|\bm{w}\|_2^2,
	}
	
	(2) logistic regression (logistic)
	\eqn{\label{eqn:logistic}}{
		\min_{\bm{w}} \frac{1}{n} \sum_{i = 1}^n \ln (1 + \exp (- y_i \bm{x}_i^T \bm{w})) + \lambda \|\bm{w}\|_2^2,
	}
	
	(3) $L2$-regularized $L1$-loss SVM (hinge)
	\eqn{\label{eqn:hinge}}{
		\min_{\bm{w}} \frac{1}{n} \sum_{i = 1}^n (1 - y_i \bm{x}_i^T \bm{w})_{+} + \lambda \|\bm{w}\|_2^2,
	}
	
	(4) $L2$-regularized $L2$-loss SVM (sqhinge)
	\eqn{\label{eqn:sqhinge}}{
		\min_{\bm{w}}\frac{1}{n}\sum_{i = 1}^n ((1 - y_i \bm{x}_i^T \bm{w})_{+})^2 + \lambda \|\bm{w}\|_2^2,
	}
	
	where $\bm{x}_i \in R^d$ and $y_i \in \{-1,+1\}$ are the feature vector and target value of the i-th example, respectively, and $\lambda > 0$ is a regularization parameter. We concatenate each row $\bm{x}_i$ of data matrix $X$ with the number 1 in (\ref{eqn:ridge}), (\ref{eqn:logistic}), (\ref{eqn:hinge}) and (\ref{eqn:sqhinge}) such that $(\bm{x}_i, 1)^T (\bm{w}, b) = \bm{x}_i^T \bm{w} + b$.

	We executed all algorithms for the binary classification on six large-scale datasets from the LIBSVM website \footnote{https://www.csie.ntu.edu.tw/~cjlin/libsvmtools/datasets/}. The information on the datasets is listed in Tab. \ref{tab:dataset}. 
	
	\tbl{\label{tab:dataset}}{Description of datasets}{lcc}{
		\hline
		Dataset & $n$ & $d$ \\
		\hline
		a9a & 32,561 & 123 \\
		covtype & 581,012 & 54 \\
		ijcnn1  & 49,990 & 22 \\ 
		w8a & 49,749 & 300 \\
		SUSY & 5,000,000 & 18 \\ 
		HIGGS & 11,000,000 & 28 \\
		\hline
	}

	\ssn{Implementations of Algorithms}
	In the preprocessing stage, each feature value for all dimensions was scaled into the range of $[-1,+1]$ by the max-min scaler. All algorithms were implemented in C++ using the armadillo linear algebra library \cite{sanderson_armadillo:_2016} and Intel MKL \footnote{https://software.intel.com/en-us/mkl}.
	
	To explore the convergence of the algorithms, we used the entire dataset to minimize the function values of the four learning models. To compare the generalization of the algorithms in classification, we randomly divided the entire dataset into three parts: 1/3 for testing, 1/5 for validation, and the remainder for training. We used the same divisions for all algorithms. After searching for the optimal parameters on the candidate set to maximize the AUC score on the validation set, we used a model corresponding to the best parameters to estimate the AUC scores on the test set.
	
	SGD, SVRG, CG and CGVR need to calculate the gradient, and S-LBFGS also calculates the Hessian matrix. In our implementations, we used numerical methods to estimate the gradient $\nabla f(\bm{x})$ and Hessian matrix $\nabla^2 f(\bm{x})$ with a small constant $\epsilon = 10^{-4}$:
	\eqna{
		\nabla f(\bm{x})_i & = & \frac{f(\bm{x} + \epsilon\bm{e}_i) - f(\bm{x} - \epsilon\bm{e}_i)}{2\epsilon}, \\
		\nabla^2 f(\bm{x})_{i,j} & = & (f(\bm{x} + \epsilon\bm{e}_i + \epsilon\bm{e}_j) 
		- f(\bm{x} + \epsilon\bm{e}_i) \nn \\
		&& - f(\bm{x} + \epsilon \bm{e}_j) + f(\bm{x}))/\epsilon^2,
	}
	where the subscript represents the i-th (or $\{i,j\}$-th) element of the matrix on the left side of the equations, and $\bm{e}_i$ is the i-th unit vector.
	
	For fair comparisons, we used the original C++ version of the LIBLINEAR solver \cite{fan_liblinear:_2008} in the discussion of generalization, which is more effective than the general LIBSVM for training SVM models on large-scale problems. We know that the measured AUC represents the area under the receiver operating characteristic (ROC) curve, which is a graphical plot that demonstrates the discrimination ability of a binary classification model when its discrimination threshold is varied. The discrimination threshold is a real value, but the output of LIBLINEAR is a discrete class label $\{-1,+1\}$. For a given threshold, many identical output values result in a large uncertainty in sorting, which was used to calculate the AUC value. Thus, we modified the predict function of LIBLINEAR to directly output the discrimination value $\bm{x}^T \bm{w} + b$. By default, LIBLINEAR optimizes the dual form of L2-regularized L2-loss SVM in the model (\ref{eqn:sqhinge}).
	
	\ssn{Parameter Investigation}
	The CGVR algorithm has two main parameters: the number of iterations of the inner loop $m$ and the number of iterations of the outer loop $T$. Parameter $T$ will be discussed in the next subsection.
	
	We selected dataset \textbf{a9a} as our research object and reported the AUC measures after 25 outer loops ($T = 25$). We used an identical random seed to initialize vector $\bm{w}_0$, where $\bm{w}_0$ is uniformly distributed over the interval $[0,1]$. In CGVR and SLBGFS, we set the sampling size $|\mathcal{S}_{k,t}|$ to $\sqrt{n}$, which was used to calculate the gradient vector and Hessian matrix of the function. For CG and CGVR, we set $c_1 = 10^{-4}$ and $c_2 = 0.1$. In SLBFGS, we set the memory size $M$ to $10$ and the Hessian update interval $L$ to $10$. SGD is accelerated in the relevant direction and dampens oscillations using the momentum method, where the momentum coefficient is commonly set to $0.9$.   For SGD, SVRG and SLBFGS, we attempted three different constant step sizes: $10^{-3}$, $10^{-4}$ and $10^{-5}$.

	\fgr{\label{fig:performance}}{Convergence on four loss functions with $\lambda = 10^{-4}$ for three variance reduction algorithms with different learning rates on a9a datasets. (a)The x-axis represents parameter $m$;(b) the y-axis is a logarithm of the value (base 10); (c) the number in the legends are the learning rate; (d) the upper and lower rows correspond to the loss and the time, respectively.}{1.1}{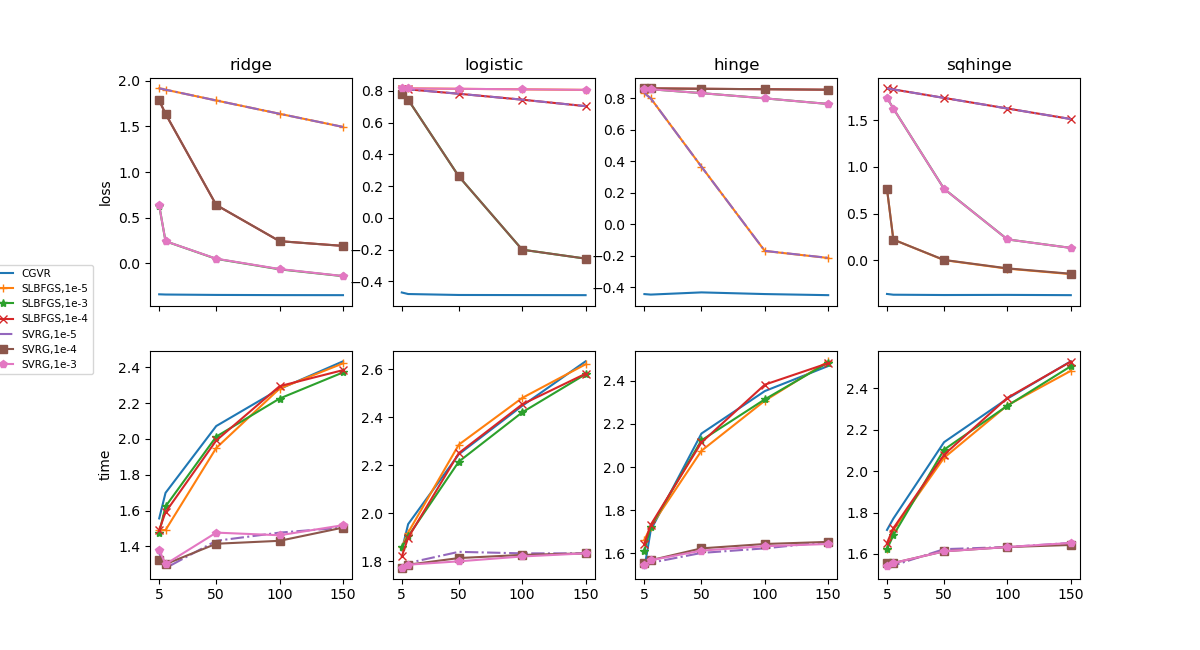}

	Fig. \ref{fig:performance} shows the function values and running time on CGVR and the other two algorithms (SLBFGS and SVRG) at different learning rates as parameter $m$ increases. The parameter $\lambda$ is set to $10^{-4}$, and the learning rate was obtained from the set $\{10^{-3},10^{-4},10^{-5}\}$. The four columns of the subfigures demonstrate that different algorithms optimize the ridge, logistic, hinge and sqhinge losses.
	
	We observe that SVRG and SLBFGS reduce the losses to some extent with increasing $m$ when setting an appropriate learning rate. The CGVR algorithm can quickly approach the minimum value of a function with only five inner loops, which slowly decreases when the $m$ value increases. Thus, CGVR is insensitive to the parameter $m$, and it requires only a few inner loops to quickly converge. 
	
	It is clear that the running time of all algorithms increase with increasing $m$, but the running time of the SLBFGS and SVRG algorithm vary with the learning rate. Although the running time of CGVR is comparable to that of the other algorithms when parameter $m$ is identical, it still has a great advantage in terms of time efficiency because CGVR only requires a few iterations of inner loops to quickly converge. 
	
	\ssn{Convergence of CGVR}
	We set the number of inner loop iterations $m$ to $50$ and compare the convergence of several algorithms on six large-scale datasets. The best model that corresponds to the optimal leaning rate was chosen for SGD, SVRG and SLBFGS, where the optimal learning rate  taken from $\{10^{-3},10^{-4},10^{-5}\}$ minimizes the value of the final iteration on the validation set.

	\fgr{\label{fig:lambda4}}{Convergence of four loss functions with $\lambda = 10^{-4}$ for five algorithms on six datasets. (a) The y-axis represents the logarithm of the loss value (base 10); (b) the x-axis represents the number of iterations of the outer loop. }{1}{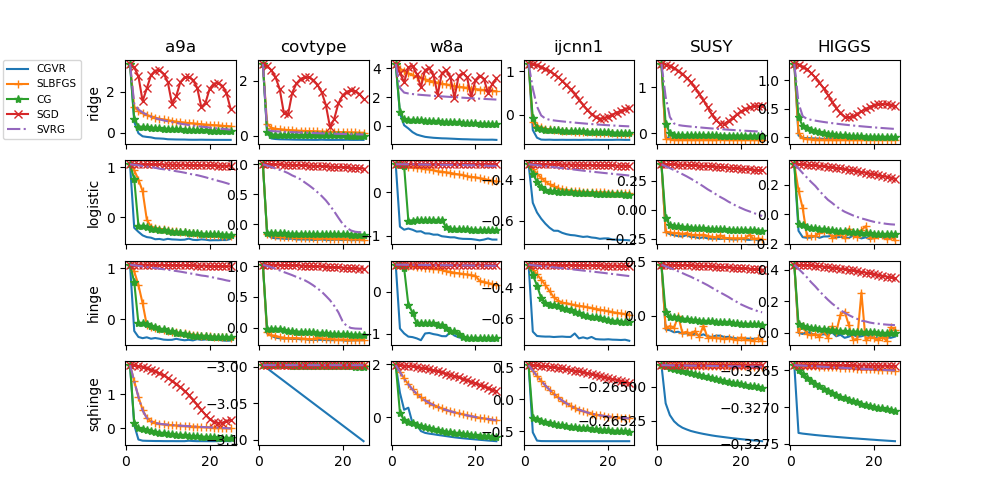}

	Fig. \ref{fig:lambda4} shows the convergence of the five algorithms with $\lambda = 10^{-4}$. 
	SGD unstably converges on the ridge model, where an inappropriate learning rate will cause large fluctuations in the loss value. SLBFGS does not show better convergence than SVRG because both SLBFGS and SVRG are sensitive to the learning rates. In general, CG converges faster than SLBFGS, SGD and SVRG. CGVR has the fastest convergence on almost all four models, even when the loss value reaches a notably small value. We also observe that all algorithms converge faster on the sqhinge model than on the other models.
	
	\ssn{Generalization of CGVR}
	To analyze the generalization of the CGVR algorithm, we show the average AUC scores of SGD, SVRG, SLBFGS, CG and CGVR on five random splits of datasets for each model in Fig. \ref{fig:alg_auc}. Furthermore, we compare the average AUC performance and execution time of CGVR and the LIBLINEAR solver on six large-scale datasets (shown in Fig. \ref{fig:cgvr-svm}). The learning rate $\alpha$ was selected from $\{10^{-3},10^{-4},10^{-5}\}$, and the regularization coefficient $\lambda$ was drawn from $\{1\times 10^{-1},5\times 10^{-2},1\times 10^{-2},8\times 10^{-3},5\times 10^{-3}\}$. The parameter $C$ in LIBLINEAR is equal to $1/(2\lambda)$. The LIBLINEAR solver will optimize the sqhinge model from (\ref{eqn:sqhinge}). The training parameters and model parameters were optimized in the space of grid $(\alpha,\lambda)$ through hold validation on the training and validation datasets.
	
	\fgr{\label{fig:alg_auc}}{Comparisons of the AUC score of CGVR and its counterpart algorithms.}{0.6}{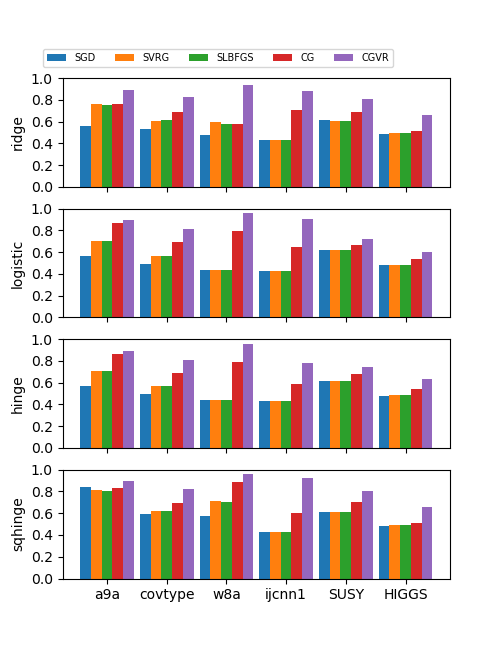}
	
	\fgr{\label{fig:cgvr-svm}}{Comparisons of the AUC score and execution time between CGVR with different models (ridge, logistic, hinge, sqhinge) and LIBLINEAR.}{0.6}{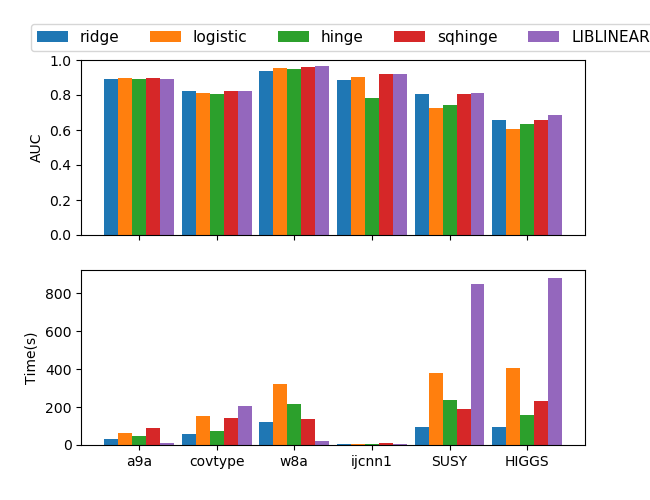}

	Fig. \ref{fig:alg_auc} shows that CGVR significantly outperforms the other counterparts on six datasets. SVRG and SLBFGS
	show close generalization performance. The classical CG algorithm shows better generalization performance than SVRG, SGD and SLBFGS. Combining the discussion on the convergence of the algorithms, we can conclude that the CGVR algorithm that achieves a smaller minimum value generally has better generalization performance in the case of appropriate regularization conditions.
	
	Fig. \ref{fig:cgvr-svm} shows that our algorithm CGVR achieves AUC scores that are comparable to those of the LIBLINEAR solver on six datasets. In the four discussed models, CGVR performs the best on all datasets when solving the sqhinge model (see \ref{eqn:sqhinge}), which is exactly the loss function optimized by the LIBLINEAR solver in default settings. Furthermore, LIBLINEAR runs faster than CGVR on small-scale datasets, such as on datasets a9a, w8a and ijcnn1, whose sample sizes are less than 100,000. However, on datasets with millions of data points, such as SUSY and HIGGS, our algorithm CGVR runs faster than LIBLINEAR, where it only iterates 25 times.

	\sn{Conclusions}
	In this paper, we proposed a new conjugate gradient algorithm based on variance reduction (CGVR). We prove the linear convergence of CGVR with Fletcher-Reeves update. The empirical results from six large-scale datasets show the power of our algorithm on four classic learning models, where these models may be convex, nonconvex or nonsmooth. The advantages of our algorithm CGVR are as follows: (1) It only requires a few iterations to quickly converge compared with its counterpart SVRG. (2) The empirical settings for CGVR always work well: the parameters are insensitive to the datasets, most of which is related to the classic Wolfe line search subroutine. (3) It requires less storage space during running, similar to the CG algorithm; it only needs to store the last gradient vector, whereas SLBFGS must store $M$ vector pairs. (4) CGVR achieves a generalization performance comparable to that of the LIBLINEAR solver for optimizing the $L2$-regularized $L2$-loss while providing a great improvement in computational efficiency in large-scale machine learning problems. 
	
	In future work, with the implementation of the algorithm using a numerical gradient, we can easily apply it to other problems, such as sparse dictionary learning and low-rank matrix approximation problems.
	
	\section*{Acknowledgments}
	This work was partially supported by the Fundamental Research Funds for the Henan Provincial Colleges and Universities in Henan University of Technology (2016RCJH06), the National Natural Science Foundation of China (61103138, 61602154, 61375039, 61473236), National Key Research \& Development Program (2016YFD0400104-5) and the National Basic Research Program of China (2012CB316301).


\begin{thebibliography}{10}
	\expandafter\ifx\csname url\endcsname\relax
	\def\url#1{\texttt{#1}}\fi
	\expandafter\ifx\csname urlprefix\endcsname\relax\def\urlprefix{URL }\fi
	\expandafter\ifx\csname href\endcsname\relax
	\def\href#1#2{#2} \def\path#1{#1}\fi
	
	\bibitem{jin_regularized_2010}
	X.-B. Jin, C.-L. Liu, X.~Hou, Regularized margin-based conditional
	log-likelihood loss for prototype learning, Pattern Recognition 43~(7) (2010)
	2428--2438.
	
	\bibitem{zhang_retargeted_2015}
	X.~Y. Zhang, L.~Wang, S.~Xiang, C.~L. Liu, Retargeted {Least} {Squares}
	{Regression} {Algorithm}, IEEE Transactions on Neural Networks and Learning
	Systems 26~(9) (2015) 2206--2213.
	
	\bibitem{jin_accelerating_2018}
	X.-B. Jin, G.-S. Xie, K.~Huang, A.~Hussain, Accelerating {Infinite} {Ensemble}
	of {Clustering} by {Pivot} {Features}, Cognitive Computation (2018) 1--9.
	
	\bibitem{jin_combination_2015}
	X.-B. Jin, G.-G. Geng, M.~Sun, D.~Zhang, Combination of multiple bipartite
	ranking for multipartite web content quality evaluation, Neurocomputing 149
	(2015) 1305--1314.
	
	\bibitem{jin_approximately_2018}
	X.-B. Jin, G.-G. Geng, G.-S. Xie, K.~Huang, Approximately optimizing {NDCG}
	using pair-wise loss, Information Sciences 453 (2018) 50--65.
	
	\bibitem{bottou_large-scale_2010}
	L.~Bottou, Large-{Scale} {Machine} {Learning} with {Stochastic} {Gradient}
	{Descent}, in: Proceedings of {COMPSTAT}'2010, 2010, pp. 177--186.
	
	\bibitem{dozat_incorporating_2016}
	T.~Dozat, Incorporating {Nesterov} {Momentum} into {Adam}, in: {ICLR}
	{Workshop}, 2016.
	
	\bibitem{reddi_convergence_2018}
	S.~J. Reddi, S.~Kale, S.~Kumar, On the {Convergence} of {Adam} and {Beyond},
	in: {ICLR}, 2018.
	
	\bibitem{kingma_adam:_2015}
	D.~P. Kingma, J.~Ba, Adam: {A} {Method} for {Stochastic} {Optimization}, Vol.
	1412, 2015.
	
	\bibitem{sutskever_importance_2013}
	I.~Sutskever, J.~Martens, G.~Dahl, G.~Hinton, On the {Importance} of
	{Initialization} and {Momentum} in {Deep} {Learning}, in: Proceedings of the
	30th {International} {Conference} on {International} {Conference} on
	{Machine} {Learning} - {Volume} 28, {ICML}'13, JMLR.org, Atlanta, GA, USA,
	2013, pp. 1139--1147.
	
	\bibitem{roux_stochastic_2012}
	N.~L. Roux, M.~Schmidt, F.~Bach, A {Stochastic} {Gradient} {Method} with an
	{Exponential} {Convergence} {Rate} for {Finite} {Training} {Sets}, in:
	Proceedings of the 25th {International} {Conference} on {Neural}
	{Information} {Processing} {Systems}, {NIPS}'12, Curran Associates Inc., USA,
	2012, pp. 2663--2671.
	
	\bibitem{shalev-shwartz_stochastic_2013}
	S.~Shalev-Shwartz, T.~Zhang, Stochastic {Dual} {Coordinate} {Ascent} {Methods}
	for {Regularized} {Loss}, J. Mach. Learn. Res. 14~(1) (2013) 567--599.
	
	\bibitem{johnson_accelerating_2013}
	R.~Johnson, T.~Zhang, Accelerating {Stochastic} {Gradient} {Descent} {Using}
	{Predictive} {Variance} {Reduction}, in: {NIPS}, {NIPS}'13, Curran Associates
	Inc., USA, 2013, pp. 315--323.
	
	\bibitem{wang_stochastic_2014}
	X.~Wang, S.~Ma, W.~Liu, Stochastic {Quasi}-{Newton} {Methods} for {Nonconvex}
	{Stochastic} {Optimization}, arXiv:1412.1196 [math].
	
	\bibitem{mokhtari_res:_2014}
	A.~Mokhtari, A.~Ribeiro, {RES}: {Regularized} {Stochastic} {BFGS} {Algorithm},
	IEEE Transactions on Signal Processing 62~(23) (2014) 6089--6104.
	
	\bibitem{moritz_linearly-convergent_2016}
	P.~Moritz, R.~Nishihara, M.~Jordan, A {Linearly}-{Convergent} {Stochastic}
	{L}-{BFGS} {Algorithm}, in: A.~Gretton, C.~C. Robert (Eds.), Proceedings of
	the 19th {International} {Conference} on {Artificial} {Intelligence} and
	{Statistics}, Vol.~51 of Proceedings of {Machine} {Learning} {Research},
	PMLR, Cadiz, Spain, 2016, pp. 249--258.
	
	\bibitem{gower_stochastic_2016}
	R.~M. Gower, D.~Goldfarb, P.~Richtárik, Stochastic {Block} {BFGS}: {Squeezing}
	{More} {Curvature} out of {Data}, 2016.
	
	\bibitem{fletcher_function_1964}
	R.~Fletcher, C.~M. Reeves, Function minimization by conjugate gradients, The
	Computer Journal 7~(2) (1964) 149--154.
	
	\bibitem{polak_note_1969}
	E.~Polak, G.~Ribiere, Note sur la convergence de méthodes de directions
	conjuguées, Revue française d'informatique et de recherche opérationnelle.
	Série rouge 3~(16) (1969) 35--43.
	
	\bibitem{hestenes_methods_1952}
	M.~R. Hestenes, E.~Stiefel, Methods of conjugate gradients for solving linear
	systems, Journal of Research of the National Bureau of Standards 49~(6)
	(1952) 409--436.
	
	\bibitem{liu_efficient_1991}
	Y.~Liu, C.~Storey, Efficient generalized conjugate gradient algorithms, part 1:
	{Theory}, Journal of Optimization Theory and Applications 69~(1) (1991)
	129--137.
	
	\bibitem{dennis_practical_1982}
	J.~Dennis, J., Practical {Methods} of {Optimization}, {Vol}. 1: {Unconstrained}
	{Optimization} ({R}. {Fletcher}), SIAM Review 24~(1) (1982) 97--98.
	
	\bibitem{dai_nonlinear_1999}
	Y.~Dai, Y.~Yuan, A {Nonlinear} {Conjugate} {Gradient} {Method} with a {Strong}
	{Global} {Convergence} {Property}, SIAM Journal on Optimization 10~(1) (1999)
	177--182.
	
	\bibitem{fan_liblinear:_2008}
	R.-E. Fan, K.-W. Chang, C.-J. Hsieh, X.-R. Wang, C.-J. Lin, {LIBLINEAR}: {A}
	{Library} for {Large} {Linear} {Classification}, J. Mach. Learn. Res. 9
	(2008) 1871--1874.
	
	\bibitem{nocedal_numerical_2006}
	J.~Nocedal, {Wright, Stephen}, Numerical {Optimization}, Springer, New York,
	2006.
	
	\bibitem{powell_restart_1977}
	M.~J.~D. Powell, Restart procedures for the conjugate gradient method,
	Mathematical Programming 12~(1) (1977) 241--254.
	
	\bibitem{al-baali_descent_1985}
	M.~Al-Baali, Descent {Property} and {Global} {Convergence} of the
	{Fletcher}—{Reeves} {Method} with {Inexact} {Line} {Search}, IMA Journal of
	Numerical Analysis 5~(1) (1985) 121--124.
	
	\bibitem{sanderson_armadillo:_2016}
	C.~Sanderson, R.~Curtin, Armadillo: a template-based {C}++ library for linear
	algebra, The Journal of Open Source Software 1.
	
\end{thebibliography}

\begin{IEEEbiography}[{\includegraphics[width=1in,height=1.25in,clip,keepaspectratio]{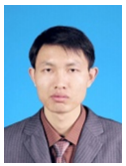}}]{Xiao-Bo Jin}
	is an Associate Professor at the School of Information Science and Engineering, Henan University of Technology. He received his Ph.D. degree in Pattern Recognition and Intelligent Systems from the National Laboratory of Pattern Recognition, Institute of Automation, Chinese Academy of Sciences, Beijing, China, in 2009. His research interests include Web Mining and Machine Learning, and his work has appeared in Pattern Recognition and Neurocomputing.
\end{IEEEbiography}

\begin{IEEEbiography}[{\includegraphics[width=1in,height=1.25in,clip,keepaspectratio]{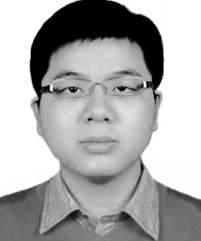}}]{Xu-Yao Zhang}
	 is an Associate Professor at National Laboratory of Pattern Recognition (NLPR), Institute of Automation, Chinese Academy of Sciences, Beijing, China. From March 2015 to March 2016, he was a visiting scholar in Montreal Institute for Learning Algorithms (MILA) at University of Montreal. His research interests include machine learning, pattern recognition, handwriting recognition, and deep learning.
\end{IEEEbiography}

\begin{IEEEbiography}[{\includegraphics[width=1in,height=1.25in,clip,keepaspectratio]{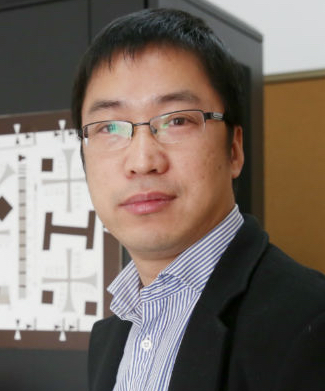}}]{Kaizhu Huang}
		works as Head of department of Electrical and Electronic Engineering in Xi’an Jiaotong-Liverpool University. Before that, he was an Associate Professor at National Laboratory of Pattern Recognition (NLPR), Institute of Automation, Chinese Academy of Sciences (CASIA). Dr. Huang obtained his Ph.D. degree from The Chinese Univ. of Hong Kong (CUHK) in 2004. He worked as a researcher  in Fujitsu R\&D Centre,  CUHK, and  University of Bristol, UK from 2004 to 2009.  Dr. Huang has been working in machine learning, pattern recognition, and neural information processing. He is the recipient of 2011 Asian Pacific Neural Network Assembly (APNNA) Distinguished Younger Researcher Award.  He has published 8 books in Springer and over 120 international research papers (45+ SCI-indexed international journals and 60+ EI conference papers) e.g., in journals (JMLR, IEEE T-PAMI, IEEE T-NNLS, IEEE T-IP, IEEE T-SMC, IEEE T-BME) and conferences (NIPS, IJCAI, SIGIR, UAI, CIKM, ICDM, ICML,ECML, CVPR).  He serves as associate editors in Neurocomputing, Cognitive Computation, and Springer Nature Big Data Analytics.   He served on the programme committees in many international conferences such as ICONIP, IJCNN, IWACI, EANN, KDIR. Especially, he serves as chairs in several major conferences or workshops, e.g., AAAI 2016-18 Senior PC, ACML 2016 Publication Chair, ICONIP 2014 (Program co-Chair), DMC 2012-2017 (Organizing co-Chair), ICDAR 2011 (Publication Chair), ACPR 2011 (Publicity Chair), ICONIP2006, 2009-2011 (Session Chair).
\end{IEEEbiography}

\begin{IEEEbiography}[{\includegraphics[width=1in,height=1.25in,clip,keepaspectratio]{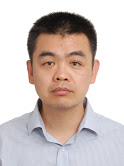}}]{Guang-Gang Geng}
	 received the Ph.D. degree from the State Key Laboratory of Management and Control for Complex Systems, Institute of Automation, Chinese Academy of Sciences, Beijing, China. He was with the Computer Network Information Center, Chinese Academy of Sciences, Beijing, in 2008. He is currently a professor with the China Internet Network Information Center. His current research interests include machine learning, adversarial information retrieval on the Web, and Web search.
\end{IEEEbiography}




\end{document}